\documentclass{article}
\usepackage{acra}

\usepackage{subfigure} 
\usepackage{epsfig} 
\usepackage{float}
\usepackage{epstopdf}
\usepackage{bm}
\usepackage{pmat}
\usepackage{amssymb} 
\usepackage{amsmath} 
\usepackage{balance, hyperref}

\title{Towards a Pantograph-based Interventional AUV for \\Under-ice Measurements}
\author{Hongkyoon Byun, Jonghyuk Kim, and Dikai Liu\\ Robotics Institute, University of Technology Sydney \\
\{hongkyoon.hugh@student,jonghyuk.kim@, dikai.liu@\}uts.edu.au \AND
Jonathan Woolfrey \\ Italian Institute of Technology, Genoa, Italy \\
jonathan.woolfrey@iit.it \\
}

\begin{document}

\maketitle

\begin{abstract}
This paper addresses the design of a novel interventional robotic platform, aiming to perform an autonomous sampling and measurement under the thin ice in the Antarctic environment. We propose a pantograph mechanism, which can effectively generate a constant interaction force to the surface during the contact, which is crucial for reliable measurements. We provide the proof-of-concept design of the pantograph with a robotic prototype with foldable actuation, and preliminary results of the pantograph mechanism and the localisation system are provided, confirming the feasibility of the system.
\end{abstract}


\section{Introduction} \label{sec:Introduction}

Autonomous underwater vehicles (AUVs) with a capability of manipulation have significant potential for industry, including environmental conservation, aquaculture, oil and gas, and infrastructure maintenance \cite{Debruyn2020medusa}. Existing working-class vehicles with manipulators (typically weighted several tons) have been successfully utilised for deep, off-shore applications, particularly in the oil and gas industry. Recently, small-scale underwater vehicles (drones) with manipulators have emerged as a new robotic technology, enabling the shallow- and cluttered-water applications, such as pylon inspection and cleaning tasks \cite{Le2020the} and underwater panel operations \cite{Cieslak2015auto}.

\begin{figure}[t]\centering \vspace{12pt}
\epsfig{figure=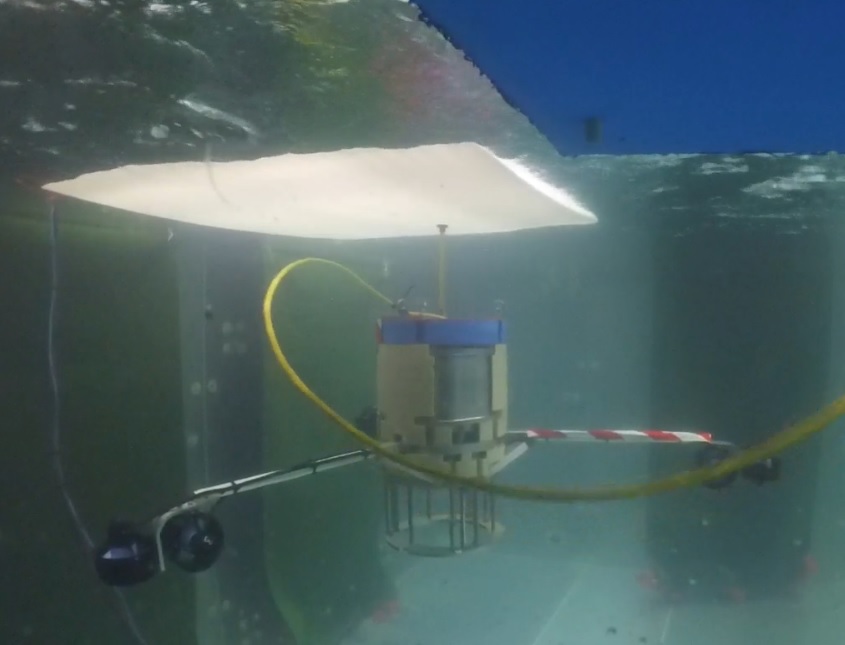,width=0.9\linewidth}
\caption{An initial (non-pantograp based) prototype of an AUV developed at the UTS for the under-ice measurement in a test water-tank environment. The AUV is equipped with foldable arms with actuators that enable deployment through a small ice-hole. A white plastic sheet is used to simulate an iceplate and a simple spring-loaded probe is used in this initial design. The tether line is used to collect optical measurements in a host system.} \label{fig:rov}
\end{figure}

Recently, it has drawn attention to using the AUVs for polar missions, such as sampling and measuring micro-algae bioactivity under thin ice plates. Comprehensive monitoring of the algae requires direct measurements, including ice thickness and photosynthesis rate from optical sensors that contact the under-ice microbial community \cite{Manes2009small}. This sensor requires sufficient contact time (e.g. one second) to acquire a reliable measurement to function correctly. To date, this sensor has been deployed by human divers or heavy articulated arms since the requirements cannot be met by existing AUV platforms. Due to fatigue, safety, cost, reachability, and time windows of operations, current approaches usually cover only a small portion of areas of interest.

\begin{figure*}[t]\centering
\epsfig{figure=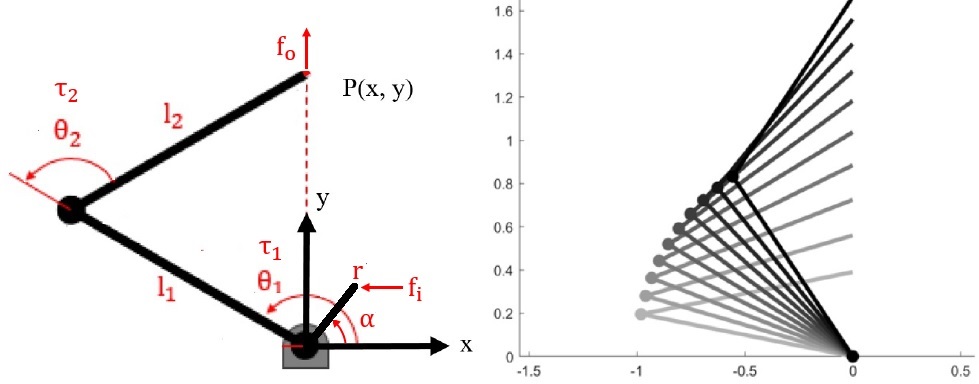,width=0.9\linewidth}
\caption{(left) The kinematic diagram of a pantograph showing two linkage bars and two joints. (right) If the pantograph constraint is enforced at the elbow joint, the end position is constrained to the vertical line.} \label{fig:1}
\end{figure*}

We provide a novel AUV system with a folding mechanism to address the challenges, which results in a compact geometry suitable for deployment through a small hole in the ice sheet. With the thrusters extended, the interventional AUV can provide sufficient control capabilities to stabilise the pose (position and orientation) against the water flow. More importantly, the AUV offers the ability of obtaining measurements by contacting the sea ice. Figure \ref{fig:rov} shows the AUV with the arm extended in a mock-up ice measurement in a water-tank experiment. A white plastic sheet was used to simulated an iceplate, and the AUV was deployed through a small radius hole ($<30$cm) simulating an ice-hole or a moonpool in an operating ship. 

This approach will enable the measurements of undersea ice photophysiology across a large enough area to extrapolate an area estimate of the photophysiological stress of the microalgal communities, such as non-photochemical quenching of photosynthesis or a decline in chlorophyll content. These phenomena are known to be early indicators of environmental changes, disease infection, and nutrient deprivation in photoautotrophs \cite{Manes2009small}. Pollutants, invasive organisms, and sudden changes in the light or temperature all can negatively impact the microalgae under the sea ice, with consequences for the rest of the ecosystem as the net primary productivity declines. This work addresses this environmental changes by developing a lightweight prototype the AUV platform suitable to collect measurements with reliable contacts to the ice surface. The initial design used a simple spring-loaded probe which, however, introduced spring deflection and nonlinear contact forces.  In this work, we investigate the use of pantograph mechanism to generate a reliabe contact force. 

The remainder of the article is outlined as follows. Section~\ref{sec:related} provides the related work in AUVs with manipulators for interventional applications, and Section \ref{sec:panto} details the kinematic model of the pantograph. Section \ref{sec:design} provides the prototype design of the AUV and pantograph, and Preliminary experimental results are provided in Section~\ref{sec:pre}, analysing the performance of the pantograph and visual navigation system. Section~\ref{sec:conclusions} will conclude with future direction.

\begin{figure*}[t]\centering
\epsfig{figure=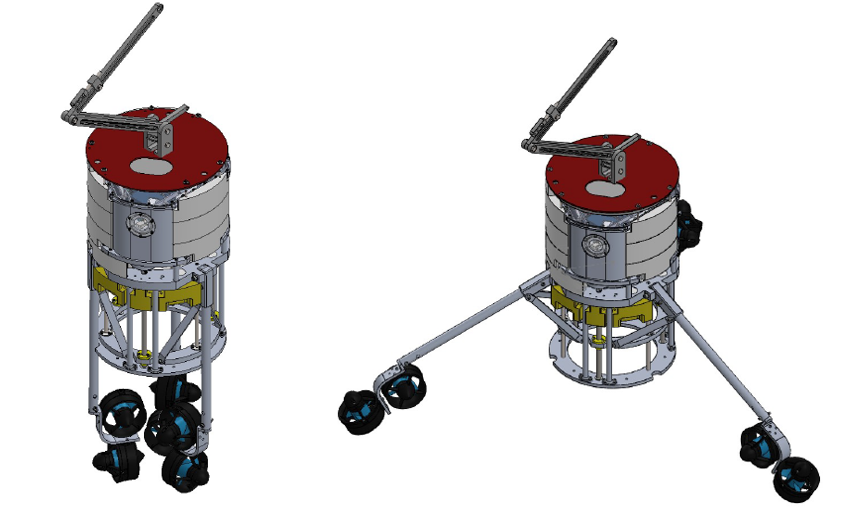,width=0.8\linewidth}
\caption{A CAD model of the pantograph-based AUV with arms folded (left), and with the arms stretched out (right). It is also equipped with a stereo-inertial system for visual localisation.} \label{fig:cad}
\end{figure*}

\section{Related Work} \label{sec:related}

AUVs have been used in many underwater applications, such as seabed survey, shipwreck search, and oil and gas industry. Such systems have been developed and studied through several projects, such as MEDUSA \cite{Debruyn2020medusa}, and PANDORA projects \cite{Maurelli2016the}. For the ice-covered ocean and polar applications, an AUV called PUMA was developed for the exploration of the Arctic seafloor \cite{Kunz2008deep}. Several underwater robots have been designed for the polar scenarios, most of them are cubic or torpedo-shaped, and some are even spherical. However, none of them can provide satisfying contact force control to obtain measurements from sensors (possibly fragile) contacting objects of interest under various uncertainties. In addition, above mentioned AUVs are not designed to operate under strong intertidal disturbances. Deployment through a small-radius ice hole or a moonpool from a ship imposes further restrictions on the design of AUVs.

\cite{Dayoub2015robotic} have demonstrated an interventional AUV in culling crown-of-thorns starfishes (COTS) in the Great Barrier Reef environment. The developed platform, COTSbot, is equipped with an injection-purpose 3-DOF manipulator underneath the vehicle, effectively shooting an end-effector to the target starfishes. \cite{Le2020the} have successfully demonstrated an underwater blast-cleaning robot, called a submersible pylon inspection robot (SPIR), at the Windang Bridge NSW, which utilises a 3-DOF manipulator for the jet-blasting. \cite{Debruyn2020medusa} demonstrated dual-robots with manipulators for underwater grasping and sampling applications. Although the underwater manipulator technology is gaining much maturity, the existing robot arms are still expensive due to the water-tight design, heavy weight due to the motors in the joints, computational complexity for Cartesian position control, and any impedence/addittance control requires additional torque/force sensors.

A practical yet effective interventional mechanism is a pantograph which has been widely applied in transportation systems. A pantograph is a redundant linkage arm that can be used to make a reliable contact between the arm and the environment. In electric trams and trains, a 'Z'-shaped pantograph has been used to provide the power, and there has been active research in maintaining a reliable contact and enough lift-forces between the pantograph and the power lines \cite{Yao2021analysis}. One of the key benefits is its simple mechanism, yet allowing the passive or active impedance/admittance control, providing a constant compliant force across a range of height variations. A manipulator with a force/torque sensor can achieve the constant force but typically requires additional cost. Another benefit is the pantograph can be driven by a passive spring or an active motor at the base, which reduces the requirements of the water-tight operation (the active servo can be housed within a vehicle body). It has also a low inertial for fast movement and no complex dynamics and control are need. 

This mechanism can also be applied for ground and aerial applications, enabling a practical solution for the compliant interaction with the controlled reaction from the contact. Unfortunately, this mechanism has not been much investigated for robotics applications, and this work study the feasibility of the pantograph system.

\begin{figure*}[th!]\centering
\epsfig{figure=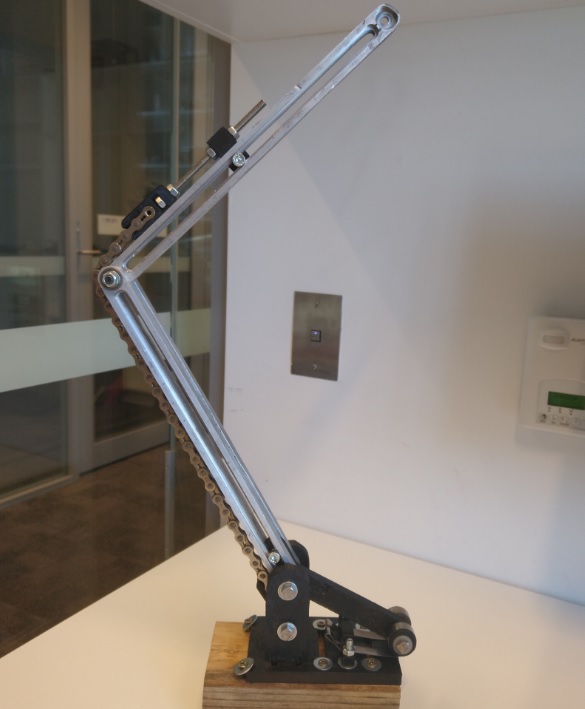,height=0.35\linewidth} \;\;\;\qquad
\epsfig{figure=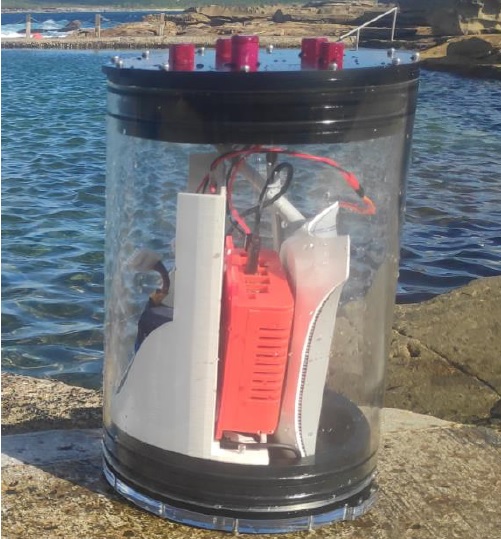,height=0.35\linewidth}
\caption{(left) A pantograph prototype (a demo link is available) and (right) a stereo-inertial system mounted in an enclosure for the test in an underwater environment (an outdoor rock pool).} \label{fig:panto}
\end{figure*}

\section{Pantograph Kinematics and Dynamics}\label{sec:panto}

Figure 2 illustrates the kinematics and geometry of a pantograph system, showing two joint angles ($\theta_1, \theta_2$) with the dimensions of linkages ($l_1, l_2$). The end position $p(x, y)$ of the pantograph then becomes,
\begin{align}\label{eq:1}
\left [ \begin{array}{l} x \\ y \end{array} \right ] =
\left [ \begin{array}{l}
l_1\cos\theta_1 + l_2\cos(\theta_1+\theta_2) \\
l_1\sin\theta_1 + l_2\sin(\theta_1+\theta_2)
\end{array} \right ],
\end{align}

The two linkage bars needs to satisfy the pantograph constraint between two bars,
\begin{align}\label{eq:constraint}
2\theta_1+\theta_2 = \pi,
\end{align}
which stems from the fact that the upper link bar must move twice as fast as the lower bar. A pair of gears with a $2:1$ ratio is used to implement this constraint. By applying this constraint, Equation (\ref{eq:1}) becomes
\begin{align}\label{eq:2}
\left [ \begin{array}{l} x \\ y \end{array} \right ] &=
\left [ \begin{array}{l}
l_1\cos\theta_1 + l_2\cos(\pi - \theta_1) \\
l_1\sin\theta_1 + l_2\sin(\pi -\theta_1)
\end{array} \right ] \\
&=\left [ \begin{array}{l}
(l_1 - l_2)\cos\theta_1 \\
(l_1 + l_2)\sin\theta_1
\end{array} \right ] \\
&=\left [ \begin{array}{c}
0 \\
2l_1\sin\theta_1
\end{array} \right ], \;\;\; \text{if}\;\; l_1 = l_2.
\end{align}
That is, by choosing the lengths of two bars equal, the motion of the endpoint can be constrained to the vertical line.

From the pantograph constraint Equation \ref{eq:constraint}, the angular velocity of two joints satisfies
\begin{align}\label{eq:3}
2\omega_1+\omega_2 = 0.
\end{align}

At the base, the acutator utilise a lever arm with the length of $r$ in the figure (perpendicular to the lower linkage bar) to which a driving input force $f_\text{i}$ applied. The resulting torques become  
\begin{align}\label{eq:decoupsys}
\tau_2 &= f_\text{o} l_2\cos(\pi-\theta_1) \\
\tau_1 &= f_\text{i} r\sin(\alpha).
\end{align}

Using the fact that $\tau_1 = 2\tau_2$ and $\theta_1-\alpha=90^\circ$, the input and output force ratio becomes
\begin{align}
\frac{f_{\text{o}}}{f_{\text{i}}} = \frac{1}{2}\frac{r\sin(\alpha)}{l_2\cos(\pi-\theta_1)} = \frac{r}{2l_2} = \text{const.}
\end{align}
showing that the output force $f_{\text{o}}$ is not affected by the configuration of the pantograph but only by the input force which can be maintained as constant from a servor motor.

\section{Design of AUV with Foldable-Arms and Pantograph}\label{sec:design}

\begin{figure*}[t]\centering
\epsfig{figure=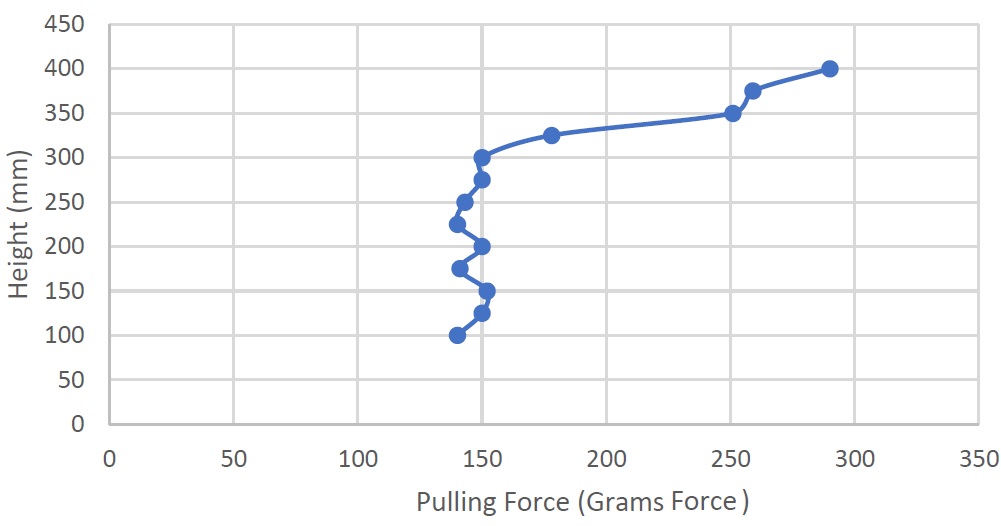,width=0.49\linewidth}
\epsfig{figure=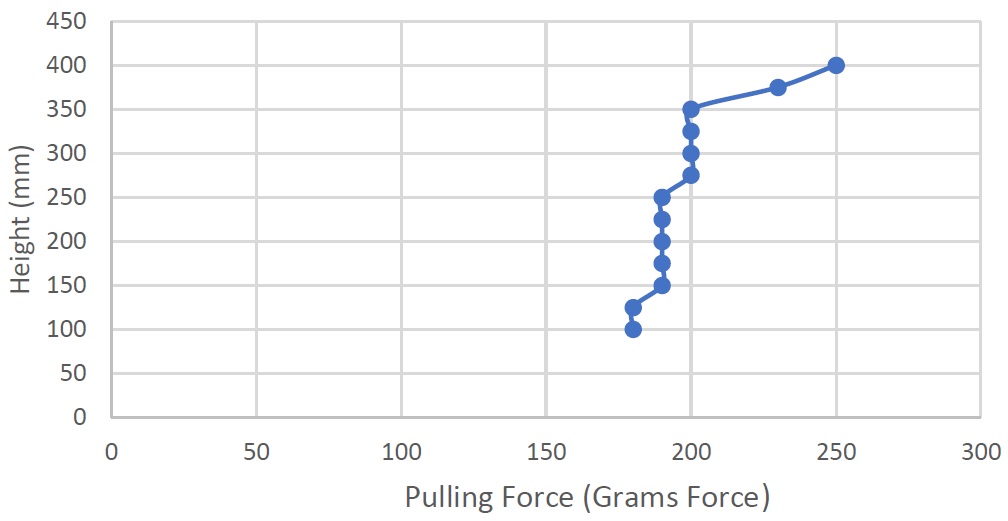,width=0.49\linewidth}
\caption{Pantograph results with the measured force at the end point with respect to the height ($y$), showing nearly-constant force measurements along the height range of $100$mm - $300$mm. The right image is obtained by using a better-quality weight sensor.} \label{fig:meas}
\end{figure*}

\begin{figure*}[th!]\centering
\epsfig{figure=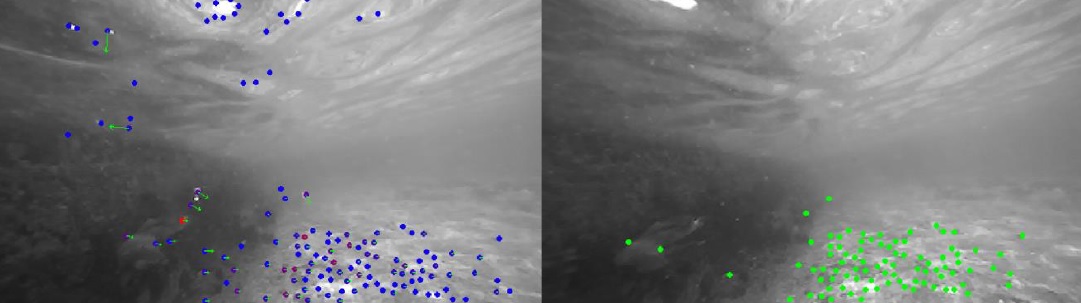,width=1\linewidth}
\caption{Feature matching results between two consecutive images from the right camera, showing the features detected (left in blue) and matched/tracked (right, in green). Most features were detected in a narrow field-of-view of the camera due to the poor visibility. } \label{fig:feature}
\end{figure*}

A custom-built underwater AUV is constructed aiming to perform the under-ice algae monitoring in an Antarctic environment. One of the critical requirements is the easy deployment of the robot from a small-radius ice hole or the moonpool of a surveyor ship. Once deployed, it should maintain a stable position with variable ocean currents and uneven ice surface, as well as contacting the surface to measure the reflectance and wavelength-dependent fluorescence emission using an optical sensing instrument on an probing arm. Figure \ref{fig:cad} shows the computer-aided design (CAD) of the robot with the folderble actuation arm mechanism. This folding configuration enables easy deployment and reliable torque control of the vehicle. The initial prototype of the robot was also shown in Figure \ref{fig:rov} in a mock-up environment, which uses a non-pantograph-based probe (a simple spring-loaded rod) and a short-base line stereo-camera which is not suitable for the localisation. 

To address these limitations, a pantograph and a visual-inertial module are developed separately, aiming for a full integration at later stage. A prototype of the pantograph is constructed using aluminium frames, which is lightweight and durable for the demonstration and experiments. To achieve the constant interaction force with adjustable force, we used a modular design: aluminium based frames for the upper and lower arms, 3D printed base frame for easy modification, a constant force spring mechanism, and a pantograph constraint coupler. The aluminium arms are accompanied with 3D printed components, which allow for more adjustments to be made to the pantograph simply by unbolting parts and moving them around the different hole positions. 

Figure \ref{fig:panto}(left) shows the constructed pantograph\footnote{A demo video (by Mr Jacob Prideaux-Remin):\href{https://youtu.be/LtjLEpsexYA}{https://youtu.be/LtjLEpsexYA}}. We chose a constant-force spring to give the pantograph mechanism a constant force along the lever arm. This is due to the fact that the constant-force spring has a minimal radius change when being pulled and uncoiled. This design is very similar to what we can find in a tape measure. The spring is coupled to a lever arm which is directly connected to the lower pantograph arms. The lever arm can be easily changed with different lengths and different arm shapes depending on the required force output.

The kinematic constraint between the joints (Equation \ref{eq:constraint}) is imposed via a chain and sprocket, inspired by the train-style constant-force pantograph and its ability to not stretch under the load that the pantograph would be experiencing. It also supports laterally, so it does not deflect away from the pantograph arms during operation. A threaded adjustment arm has been used to adjust the pantograph with precision to achieve a linear movement by adjusting the chain length along the arm. Figure \ref{fig:panto}(right) shows the enclosure of the visual-inertial localisation system for the purpose of testing in an outdoor underwater environment. The localisation system utilises a stereo-inertial camera and an embedded computer in a water-tight enclosure. A VINS-fusion ROS package is used to test the performance in an outdoor rock pool environment.

\section{Preliminary Results}\label{sec:pre}

Due to the limited access to the water-tank facility at the UTS, we only presents the individual test results of the pantograph and localisation system in this work, although we are working towards the full integration and testing.  

\begin{figure*}[th]\centering
\epsfig{figure=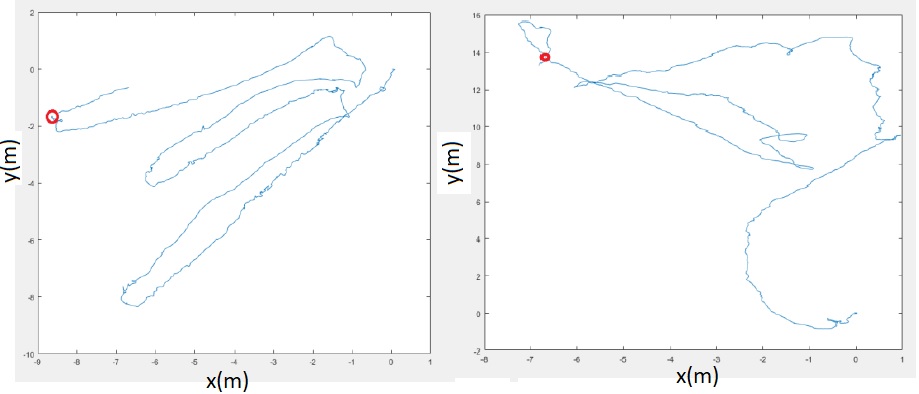,width=0.8\linewidth}
\caption{(left) Preliminary results from the VINS fusion using the ZED2 camera. (right) Odometry output from the ZED2 camera showing significant drift in the position. The ground truth was not available for accurate evaluation, but the estimated trajectory properly captured the sweeping lawn-mower pattern during the experiments.} \label{fig:vins}
\end{figure*}

The designed pantograph is tested in the laboratory to analyse the force-height relationship. The pantograph mechanism is set at $100$mm between the lower axis and top of the pantograph arm, and the force is measured using a simple scale. It is then extended vertically by $25$mm, and then the force is measured, which process is repeated until $400$mm height is reached. Figure \ref{fig:meas} shows two test results of the force-height measurements. The first test (left) shows a relatively constant force between $100$ mm and $300$ mm while showing stong nonlinear effects when the height exceeds $300$mm. The second test (right) with a different measuring device is able to achieve a relatively constant force output between $100$mm and $350$mm. The variations of results stem from several factors, such as the friction in the system, which can cause resistance against the spring mechanism at certain points. Additionally, there is friction between the rope used and the pulley, which could cause the pantograph to exert all its force as the string is holding it back. The measuring device used is also a low-cost one causing some inaccuracy. However, the experiments clearly demonstrate the feasibility of the pantograph achieving a nearly constant force at the endpoint over a range of height variations.

A stereo-inertial navigation system is assembled and tested in a realistic underwater environment, an outdoor rock pool, as shown in Figure \ref{fig:panto} (right), which consists of a ZED2 camera and an NVIDIA NX embedded computer in a water-tight enclosure. The water was relatively clear with natural features in the rock wall and floor. The camera system was pointed forward and followed a lawn-mower trajectory pattern. Figure \ref{fig:feature} (left) shows the features detected from the right camera, and (right) the tracked feature (right) in the subsequent image. It can be observed that the detection range is quite narrow and short due to the low illumination and blurry features. Figure \ref{fig:vins} shows the preliminary localisation results using the VINS fusion package (left) and the direct ZED2 odometry output (right). It can be clearly seen that the ZED2 odometry output (IMU and stereo) is drifting rapidly, while the VINS fusion method shows more consistent accuracy. No ground truth trajectory is recorded in this preliminary experiment, but the sensing system followed a lawn-mower sweeping pattern which can be seen in the estimated trajectory. From the post-analysis, the narrow field-of-detection caused very sparse loop-closures between the tracks, distorting the straight trajectory. We are planning further experiments with a trajectory that has enough overlapping between the neighbouring routes and testing different viewing angles of the camera to maximise the chance of loop-closures.

\section{Conclusions}\label{sec:conclusions}
We presented a novel interventional AUV system that incorporates a pantograph-based interventional mechanism and a folding actuation. We reviewed the kinematics and dynamics of the pantograph system and provided the prototypes of the AUV platform and the pantograph system. Preliminary experimental results confirms that the system can maintain a constant contact force of around $190$ gram-force over the height variations of $100$mm - $300$mm, sufficient for the purpose of the optical measurements. The visual-inertial localisation system also showed consistent localisation performance in the realistic underwater condition. The future direction is to integrate the pantograph mechanism on the AUV and the autonomous operation of the robot using the visual-inertial localisation outputs.

\section*{Acknowledgments}
The prototype of the pantograph was constructed as part of a student design project at the UTS: Mr Jacob Prideaux-Remin, Mr Van-Troung Lam, Mr Darren Hayes, Mr Abdullah Azam, and the navigation system was tested by Mr Aman Singh. The initial AUV control system was designed by Dr Wenjie Lu at the UTS.


\balance

\end{document}